\documentclass{article}

\usepackage[preprint]{neurips_2025}

\usepackage[utf8]{inputenc} 
\usepackage[T1]{fontenc}    
\usepackage{hyperref}       
\usepackage{url}            
\usepackage{booktabs}       
\usepackage{amsfonts}       
\usepackage{nicefrac}       
\usepackage{microtype}      
\usepackage{xcolor}         
\usepackage{graphicx}
\usepackage{amsmath}
\usepackage{subcaption}
\usepackage{amsthm}
\usepackage{tabularx}
\usepackage{wrapfig}

\title{Global to Local: Topology-Preserving Adaptive Graph Pooling via Granular-Ball}

\author{%
	Sen Zhao\\
	\And
	Gaojie Xu\\
	\And
	Shuyin Xia*\\
	\And
	Yifan Guan\\
	\And
	Yi Liu\\
	\And
	Yi Wang\\
	\And
	Wei Wang
}

\begin{document}

\maketitle

\begin{abstract}
Graph pooling aims to compress the graph, including both node embeddings and their underlying topological patterns, into a more compact representation. Previous works focus primarily on the overly fine-grained representation of nodes, progressively coarsening the graph by removing nodes or merging them into clusters, thus neglecting the global-to-local patterns and adaptive granularity of the graph’s topological structure. 
In the real scenario, graphs as a whole can
be considered the coarsest level of granularity, encapsulating the
global topological structure, with progressively finer-grained local
topological structures represented from top to bottom. This process continues until the adaptive granularity for each subdomain is
reached. To this end, we propose a novel Topology-Preserving Adaptive
Graph Pooling (TPAGP) method that dynamically partitions graphs into granular balls by integrating node features and topological information, enabling the generation of multi-granularity representations that effectively capture both local and global structural patterns. Additionally, we design a multi-granularity graph network model that facilitates feature interaction and optimization across different granularities, significantly enhancing performance in graph classification tasks. Experimental results demonstrate that TPAGP outperforms existing pooling methods across various benchmark datasets, effectively mitigating information loss caused by fixed-granularity strategies. The code is available at \url{https://anonymous.4open.science/r/TPAGP}.
\end{abstract}

\section{Introduction}
Graph Neural Networks (GNNs) have become essential tools for processing graph-structured data due to their ability to propagate and aggregate node features along edges~\cite{kipf2017semi,xu2018powerful}. While GNNs effectively learn node-level representations, many downstream tasks, such as graph classification~\cite{wu2020comprehensive,wu2022graph}, require summarizing entire graphs into compact representations. Graph pooling plays a crucial role in generating such graph-level representations by compressing node embeddings and their underlying topological patterns into a unified format. 

Early works~\cite{atwood2016diffusion,xu2018powerful} primarily employed global pooling methods, which aggregate all node embeddings using simple operations like summation or averaging. However, these methods often fail to preserve the graph’s topological structure, as they neglect finer-grained structural patterns within the graph. 
\begin{figure}[t]
  \centering
  \includegraphics[width=0.99\textwidth]{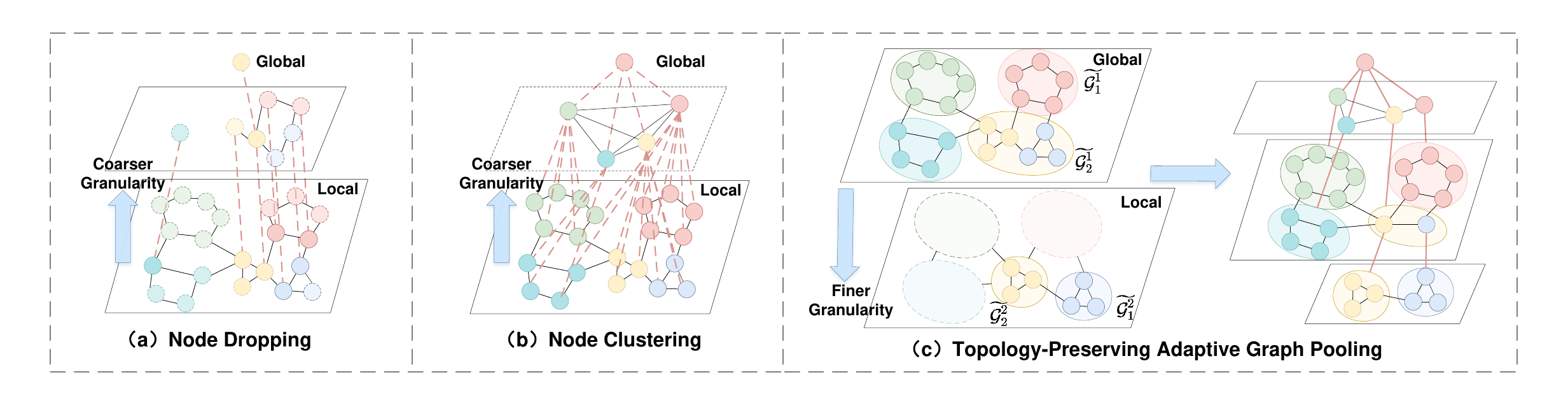}
  \caption{Comparison of hierarchical graph pooling methods.}
  \label{fig:motivation}
\end{figure}
To overcome these limitations, researchers have introduced hierarchical pooling methods~\cite{gao2019graph,lee2019self,ying2018hierarchical}, which iteratively coarsen graphs by merging nodes or subgraphs. These approaches are designed to capture multi-scale structural information, offering a more topology-aware representation of graphs. Broadly, hierarchical pooling methods can be classified into two categories: node dropping and node clustering. 
Node dropping methods (as illustrated in Fig.~\ref{fig:motivation}a) reduce the graph size by selecting a subset of nodes based on their importance scores, discarding less important nodes during the pooling process~\cite{gao2019graph,lee2019self,ranjan2020asap,ying2024boosting}. While these methods offer advantages in memory efficiency and computational scalability, the removal of nodes can compromise the graph's topological structure and result in information loss~\cite{bianchi2023expressive}. 

Node clustering methods (as illustrated in Fig.~\ref{fig:motivation}b), in contrast, focus on preserving topological information by learning a cluster assignment matrix, which groups nodes into clusters and progressively aggregates them into coarser graph representations~\cite{ying2018hierarchical,bianchi2020spectral,DBLP:conf/iclr/Song0WZL24}. Despite their effectiveness, both node dropping and node clustering methods focus primarily on local graph information, progressively coarsening the graph by removing nodes or merging them into clusters, thus neglecting the global-to-local patterns and adaptive granularity of the graph’s topological structure.

In real-world graph pooling, graphs contain subregions with varying granularities and distinct topological structures. Nodes within the same subregion tend to be more similar, indicating local topological coherence. 

As shown in Fig.~\ref{fig:motivation}(c), the entire graph forms the coarsest granularity, while finer structures emerge hierarchically until each subdomain reaches its adaptive level.
In the example shown in Fig.\ref{fig:motivation} (c), $\widetilde{\mathcal{G}_1^1}$ corresponds to an appropriate level of granularity, while $\widetilde{\mathcal{G}_2^1}$ should be further decomposed into $\widetilde{\mathcal{G}_1^2}$ and $\widetilde{\mathcal{G}_2^2}$ at a finer granularity, as the structural coherence and node similarity within $\widetilde{\mathcal{G}_2^1}$ indicate that its subdomains could be further refined to capture more detailed topological patterns. Previous works, as illustrated in Fig.~\ref{fig:motivation} (a), (b), overlook the global-to-local topological patterns by starting with an overly fine-grained representation of individual nodes and recursively coarsening the graph to capture topological information from the local to the global scale. This approach shifts attention to global topological relationships only after simplifying the local details, which can disrupt the graph’s original topology and hinder the ability to capture global structural patterns. Furthermore, pooling the graph from the finest granularity disregards the adaptive nature of granularity in different topological subdomains, limiting the method’s capacity to capture the full range of structural complexities across the graph.

Actually, preserving the graph's topology from global to local scales with adaptive granularity is non-trivial, due to two challenges: 1) Topological Structure Refinement: Topological structures in a graph are characterized by nodes with high connectivity and strong feature similarity. These structures should be modeled at the appropriate granularity, ensuring that neither coarser-grained parent graphs nor finer-grained subgraphs exhibit a higher degree of topological coherence than the granularity at which the structure is being analyzed.
Inspired by the advantages of granular-ball computing \cite{xia2019granular,xie2023efficient} in modeling multi-granular characteristics for scattered data, we explore its potential for graph refinement and propose an Adaptive Granularity graph refinement mechanism to effectively capture and preserve the topological structures within the graph. 2) Pooling of Topological Structures: The second challenge lies in how to pool these topological structures, especially considering the tree-like structure formed through graph refinement, which progresses from global to local. Due to the adaptive granularity, the depth of branches may vary across different subdomains, leading to inconsistent structural levels. This necessitates a pooling mechanism capable of handling varying depths and granularities, preserving both local and global topological patterns while adapting to the inherent complexity introduced by the hierarchical refinement process. 

To this end, we propose a novel Topology-Preserving Adaptive Graph Pooling (TPAGP) method, inspired by the multi-granularity application of granular-ball computation to discrete data points. TPAGP generates multiple granular-balls, each representing a portion of the graph structure, by adaptively partitioning the graph and integrating node features with topological information. This partitioning enables the creation of subgraphs at different granularities, capturing both local and global structural patterns effectively. The pooling process is dynamic and data-driven, adjusting the granularity based on the characteristics and features of the graph, ensuring that important node relationships and attributes are preserved. Additionally, TPAGP integrates a multi-granularity graph network model, where features from different granular-balls are processed and optimized collaboratively, facilitating efficient information transfer across granularities. This enables the model to learn a more comprehensive graph representation by capturing interactions between features at various granular levels, ultimately improving performance in graph-based tasks. The adaptive nature of TPAGP allows it to flexibly handle heterogeneous graph data, making it particularly effective for complex graph structures.

\section{Relation work}

In graph neural networks, graph pooling is a technique that downscales the graph by aggregating or selecting nodes, helping to learn graph-level representations for tasks like classification. 
Early works~\cite{atwood2016diffusion,simonovsky2017dynamic,xu2018powerful} primarily employed global pooling methods that learn a graph-level representation by aggregating the features of all nodes. Common global pooling methods include summing or averaging all node representations \cite{Set2Set}, as well as using Long Short-Term Memory  networks to aggregate node features \cite{hochreiter1997long}. However, the main limitation of these methods is their inability to fully capture the intricate relationships or structural dependencies between nodes, as they often treat node features independently without considering the graph's underlying topology. To address this, Zhang et al. \cite{DGCNN} proposed the DGCNN model, which introduces the SortPooling layer to reorder node feature descriptors based on their structural properties, better preserving the graph's topology. While these methods are straightforward and effective, they often fail to capture the hierarchical or multi-scale structure that is frequently present in real-world graphs.

Hierarchical pooling methods progressively reduce the size of the graph by mapping it to coarser versions, enabling the learning of higher-level representations. These methods can be broadly categorized into node dropping and node clustering. Node-dropping methods assess the importance of nodes and select a subset to form a new, smaller graph. For example, TopKPool \cite{gao2019graph} learns a learnable projection vector to score nodes and selects the highest-scoring nodes based on a predefined pooling ratio. SAGPool \cite{lee2019self} uses Graph Neural Networks to incorporate graph structure when scoring nodes. 
TIP~\cite{ying2024boosting} further advances this approach by integrating topological information and feature preservation into the node dropping process, ensuring that both local and global graph structures are effectively captured at different levels of granularity. 
On the other hand, node-clustering methods group nodes into clusters, reducing the graph's size while preserving its structural integrity. For instance, DiffPool \cite{ying2018hierarchical} learns a node clustering assignment matrix to aggregate nodes into a new subgraph, while MinCutPool \cite{bianchi2020spectral} optimizes the clustering assignment with a min-cut objective. Although node-clustering methods are effective in preserving the graph’s structural information, the reliance on fixed compression quotas may still result in the loss of important local structures.  GPN~\cite{DBLP:conf/iclr/Song0WZL24} further refines this process by adaptively learning a personalized pooling structure for each graph. Inspired by bottom-up grammar induction. Previous works focus primarily on the overly fine-grained representation of nodes, progressively coarsening the graph by removing nodes or merging them into clusters, thus neglecting the global-to-local patterns and adaptive granularity of the graph’s topological structure.

\section{Our Method}
We proposed Topology-Preserving Adaptive Graph Pooling via Granular-Ball(TPAGP). 

\subsection{ Topology-Preserving Adaptive Graph Pooling via}

Topology-Preserving Adaptive Graph Pooling process consists of two main stages: \textbf{Global Topology-Aware Granular Initialization} and \textbf{Local Topology-Preserving Granular Optimization}. 

\begin{figure*}
  \centering
  \includegraphics[width=1\textwidth]{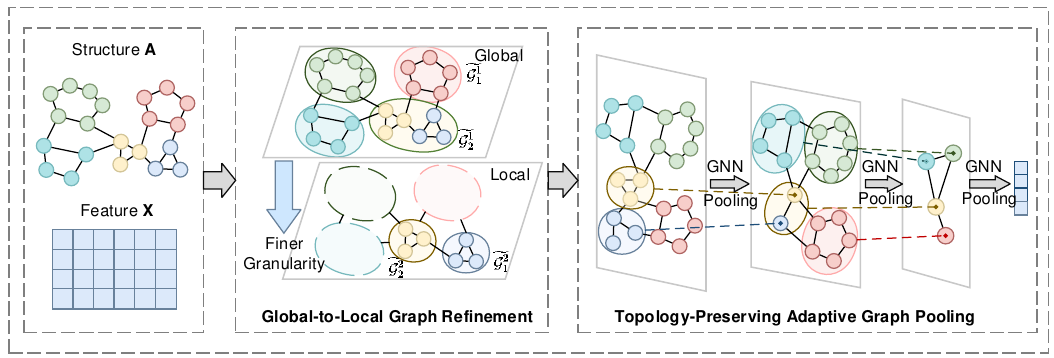}
  \caption{The TPAGP framework utilizes Granular-Ball Graph Pooling to adaptively optimize and hierarchically pool graph structures, followed by GNN for feature learning. $|G|$ represents the number of nodes.}
  \label{framework}
\end{figure*}

In the \textbf{Global Topology-Aware Granular Initialization} stage, the input graph $G = (V, E)$ is initially considered as a single coarse-grained granular-ball, where all nodes are grouped into one cluster, following the cognitive principle of "global precedence". This initialization reflects a topologically holistic view, abstracting the graph as an indivisible whole. Subsequently, this granular-ball is split into $\sqrt{N}$ granular-balls \cite{yu2001upper}, where $N$ is the number of nodes in the graph. The goal of this process is to preserve the intrinsic topological characteristics while achieving an initial coarse granularity.

To split the graph \( G \) into \( \sqrt{N} \) granular-balls, the process is as follows:

\textbf{Identifying Centers:}  
To ensure that the initial partition respects the graph’s topology, we select the set of $\sqrt{N}$ nodes with the highest degrees as the centers, since high-degree nodes are often topologically influential and well-connected hubs. Let $C$ denote the set of initial centers:
\begin{align}
    &C = \{v_1, v_2, \dots, v_{\sqrt{N}}\}, \nonumber \\ 
    &\text{where } \deg(v_i) \geq \deg(v_j) \text{ for } v_i \in C, v_j \notin C.
\end{align}
This selection encourages the initial granular-balls to be centered around structurally important nodes, promoting topological representativeness.
    
\textbf{Multi-Source BFS Assignment:  }
To assign nodes to granular-balls, we perform a multi-source breadth-first search (BFS) from all centers in $C$ simultaneously. Each node $u \in V$ is assigned to the granular-ball $GB_i$ corresponding to its closest center $v_i \in C$ based on the shortest path in the graph:
\begin{align}
GB(u) = \arg\min_{v_i \in C} d(u, v_i),
\end{align}
where $d(u, v_i)$ is the shortest path distance from node $u$ to center $v_i$. This BFS-based propagation respects the underlying graph topology by following edge connectivity patterns and path structures. It ensures that each granular-ball forms a topologically cohesive region, preserving local neighborhood structures and community-like formations.

In In the \textbf{Local Topology-Preserving Granular Optimization} stage, each granular-ball \( GB_i \) is evaluated to determine whether its local structural quality can be improved by further partitioning. The splitting criterion is based on a comparison of granular-ball quality, which reflects both intra-ball cohesion and structural consistency. A granular-ball is split into two child granular-balls if the parent’s quality \( Q_{GB_i} \) is lower than the combined quality of its two potential children \( Q_{GB_j} + Q_{GB_k} \), where \( GB_j \) and \( GB_k \) are the children of \( GB_i \):
\begin{align}
    &Q_{GB_i} < Q_{GB_j} + Q_{GB_k}, \nonumber \\
    &\text{where } GB_j, GB_k \in \text{children of } GB_i.
\end{align}

To perform the split in a topology-preserving manner, the two nodes with the highest degrees within \( GB_i \) are selected as the new centers, under the assumption that high-degree nodes are topological anchors within the local subgraph. Then, a multi-source BFS is executed to assign each node in \( GB_i \) to the closest center, forming two topologically cohesive child granular-balls. The BFS ensures that the assignment respects edge connectivity and shortest-path distances within the subgraph, preserving the underlying local topology.

This recursive splitting process continues until all granular-balls satisfy the condition:
\begin{align}
    &Q_{GB_i} \geq Q_{GB_j} + Q_{GB_k}, \nonumber \\
    &\text{where } GB_j, GB_k \in \text{children of } GB_i.
\end{align}

\subsection{Adaptive Granularity Quality Evaluation Based on Topological Data Analysis}

To quantify the performance of the granular-ball method in graph data processing and evaluate its adaptive granularity selection capability, we refer to a quality evaluation method based on Topological Data Analysis (TDA) \cite{chen2023topological}. This method leverages persistent homology to analyze the topological features of graph structures and quantifies the effectiveness of granular-ball methods through the distribution of persistence bar durations across multiple dimensions.

From a topological perspective, we aim to evaluate the quality of granular-ball splitting, as this directly influences the effectiveness of downstream tasks such as graph pooling. A good splitting scheme should preserve important topological structures in the data, avoiding over-segmentation that destroys global patterns, or under-segmentation that overlooks local features. Therefore, assessing the impact of granularity adaptation through TDA provides a theoretical and interpretable foundation for understanding how well granular-balls capture essential graph characteristics.

In the granular-ball method, adaptive granularity is achieved by dynamically adjusting the size and center distribution of balls to capture both global and local features of the graph structure. Given a graph \( G = (V, E) \), where \( V = \{v_1, v_2, \dots, v_n\} \) is the set of nodes, each node \( v_i \) is associated with a feature vector \( \mathbf{f}_i \in \mathbb{R}^d \). The feature vectors are combined into a point cloud \( P = \{\mathbf{f}_1, \mathbf{f}_2, \dots, \mathbf{f}_n\} \subset \mathbb{R}^d \), which serves as the basis for subsequent topological analysis.

Based on the point cloud \( P \), we construct a Rips complex \( \text{Rips}(P, \epsilon) \) by setting a distance threshold \( \epsilon > 0 \). A \( k \)-simplex \( \sigma \) in the Rips complex is defined as:  
\begin{align}
\sigma = \{v_0, v_1, \dots, v_k\} \subset V, \quad \text{where} \quad \forall u, v \in \sigma, \, \text{dist}(\mathbf{f}_u, \mathbf{f}_v) \leq \epsilon,
\end{align}
where \( \text{dist}(\cdot, \cdot) \) denotes the Euclidean distance. By adjusting \( \epsilon \), the Rips complex can flexibly capture the features of the point cloud at different granularities, reflecting the performance of the granular-ball method at various scales.

The Rips complex is stored using a simplex tree, and persistent homology is computed to extract persistence barcodes across different dimensions. Each persistence bar \( [b_{i,j}, d_{i,j}) \) corresponds to an \( i \)-dimensional homology class, where \( b_{i,j} \) and \( d_{i,j} \) represent the birth and death times of the homology class, respectively. The duration of a persistence bar is defined as:  
\begin{align}
\text{duration}_{i,j} = d_{i,j} - b_{i,j}.
\end{align}
To eliminate the influence of infinite values, only bars with finite death times are retained. The duration is adjusted based on the evaluation mode:
\begin{itemize}
    \item \textbf{Superlevel set mode:} The duration is negated:  
    \begin{align}
    \text{duration}_{i,j} = -(d_{i,j} - b_{i,j}).
    \end{align}
    \item \textbf{Sublevel set mode:} The duration remains positive:  
    \begin{align}
    \text{duration}_{i,j} = d_{i,j} - b_{i,j}.
    \end{align}
\end{itemize}

To comprehensively evaluate the adaptive granularity of the granular-ball method, we sum the durations of persistence bars across all dimensions. The granular-ball quality \( Q \) is defined as:
\begin{align}
Q = \sum_{i=0}^{\text{dim}} \sum_{j=1}^{n_i} \text{duration}_{i,j},
\end{align}
where \( n_i \) is the number of persistence bars in dimension \( i \). For specific dimensions \( \text{dim} \), the formula expands as follows:

\begin{align}
For\text{ dim} = 0: & 
Q = \sum_{j=1}^{n_0} (d_{0,j} - b_{0,j}), \\
For\text{ dim} = 1: &
Q = \sum_{j=1}^{n_0} (d_{0,j} - b_{0,j}) + \sum_{k=1}^{n_1} (d_{1,k} - b_{1,k}), \\
For\text{ dim} = 2: &
Q = \sum_{j=1}^{n_0} (d_{0,j} - b_{0,j}) + \sum_{k=1}^{n_1} (d_{1,k} - b_{1,k}) + \sum_{l=1}^{n_2} (d_{2,l} - b_{2,l}).
\end{align}

This formula quantifies the ability of the granular-ball method to capture graph structures at multiple scales, especially under adaptive granularity adjustments. By integrating TDA and persistent homology, this method dynamically adapts to different granularities of graph structures, providing a comprehensive evaluation of the granular-ball method's ability to capture both global and local features. This adaptive granularity evaluation method serves as a critical basis for optimizing and applying granular-ball techniques in graph pooling and other graph representation learning tasks.

\subsection{Model Architecture}

The proposed model integrates hierarchical graph convolution and topology-preserving adaptive pooling to achieve multi-level representation learning for graph classification tasks. In particular, the model further incorporates granular-ball graph pooling, where the hierarchical structure is constructed based on granular-ball generation, enabling a bottom-up learning process from local regions to global structures.

The architecture consists of alternating graph convolution and pooling operations, where the graph structure is progressively coarsened. Different from conventional pooling methods, the proposed model first transforms the original graph into a hierarchical granular-ball structure. Specifically, the original graph is partitioned into multiple granular-balls, each representing a subgraph of structurally similar nodes. These granular-balls naturally form a tree-like hierarchy from fine-grained local regions to coarse-grained global representations.

Based on this structure, the model performs graph neural network learning in a bottom-up manner. At the initial stage, graph convolution is applied to the original graph to aggregate neighborhood information and update node representations. The feature propagation at each layer is formulated as:
\begin{equation}
h_i^{(l+1)} = \sigma \left( \sum_{j \in \mathcal{N}(i)} \frac{1}{\sqrt{|\mathcal{N}(i)| |\mathcal{N}(j)|}} W^{(l)} h_j^{(l)} + b^{(l)} \right),
\end{equation}
where $h_i^{(l)}$ represents the feature of node $i$ at layer $l$, $\mathcal{N}(i)$ denotes the set of neighboring nodes of node $i$, and $W^{(l)}, b^{(l)}$ are learnable parameters. The activation function $\sigma(\cdot)$ introduces non-linearity.

After obtaining node embeddings, granular-ball graph pooling is performed. At the lowest level, each granular-ball contains a subset of original nodes, and a mixed graph is constructed where both original nodes and granular-ball nodes coexist. Graph convolution and pooling are then jointly applied to this graph:
\begin{itemize}
\item Compute node embeddings using the first-stage graph convolution network $h_{\text{gnn1}} = \text{GNN}(h, A)$.
\item Apply topology-preserving adaptive granular-ball pooling (TPAGP) to generate a coarser graph $G'$.
\item In the pooled graph, nodes within the same granular-ball are aggregated into a single super-node.
\end{itemize}

In the next stage, the granular-balls obtained from the previous level are treated as new nodes. A new graph is constructed where these higher-level granular-ball nodes coexist with remaining nodes. Graph convolution is then applied again to this updated graph, followed by another round of pooling:
\begin{itemize}
\item Recompute node embeddings on the updated graph $h_{\text{gnn2}} = \text{GNN}(h', A')$.
\item Perform granular-ball pooling to further merge nodes into higher-level granular-balls.
\end{itemize}

This process is repeated iteratively, where granular-balls at lower levels are progressively merged into higher-level granular-balls, forming a hierarchical representation from local to global structures. The pooling process terminates when all nodes in the graph are represented by granular-balls, or when the graph structure remains unchanged between successive iterations, ensuring adaptive granularity.

Finally, for graph classification, a multiset readout function is applied to the final coarsened graph to obtain a graph-level embedding. The resulting representation is then fed into a multi-layer perceptron (MLP) for prediction.

The entire model is trained end-to-end using backpropagation. By integrating granular-ball graph pooling with hierarchical graph convolution, the proposed architecture effectively preserves structural information at different granularities while enabling scalable and expressive graph representation learning.
\section{Experiment}

In the following section, we present a comprehensive set of experiments designed to evaluate the effectiveness of our TPAGP. These experiments aim to address the subsequent research questions:
\textbf{RQ1:} How does the performance of TPAGP compare to current state-of-the-art methods?
\textbf{RQ2:} What is the impact of each key component on the overall performance of TPAGP?
\textbf{RQ3:} How do different hyper-parameter settings affect TPAGP's performance?
\textbf{RQ4:} How does TPAGP perform across various case study scenarios, and what insights can be gained from these applications?

\subsection{Datasets}
Graph classification is evaluated using six standard graph datasets \cite{Morris+2020}, which encompass various fields, including biology, chemistry, social network, and medicine, as detailed in Table \ref{tab:dataset}. In the table, Size denotes the number of graphs in the dataset, \( N \) denotes the average number of nodes, and \( M \) denotes the average number of edges. These datasets are widely used in graph learning and graph neural network research across various domains, refer to the appendix \ref{appendix:dataset} for specific introductions.

\subsection{Experimental Settings}
All experimental results are obtained from the on a Linux server equipped with one Tesla V100 GPU (32GB memory). The software environment consists of Python 3.9.21, PyTorch 1.11.0, and CUDA 11.3. The key hyperparameter settings for each dataset can refer to the appendix \ref{appendix:Parameter}.

\begin{table}[htbp]
  \caption{Dataset Information.}
  \centering
  \label{tab:dataset}
  \begin{tabular}{cccc}
    \toprule
    Dataset&Size&N&M\\
    \midrule
MUTAG  & 188 & 17.93 & 19.79 \\
MSRC\_9  & 221 & 40.58 & 97.94 \\
BZR & 405 & 35.75 & 38.36 \\
DD & 1178 & 284.32 & 715.66\\
PTC\_MR  & 344 & 14.29 & 14.69 \\
IMDB-MULTI(IMDB)  & 1500 & 13.00 & 65.94 \\
  \bottomrule
\end{tabular}
\end{table}

\subsection{Comparison Experiments}

To answer \textbf{RQ1}, we compare the proposed TPAGP framework with various baseline methods. They are \textbf{(\uppercase\expandafter{\romannumeral1}) GNN-based methods}: GCN \cite{kipf2017semi}, GAT \cite{velivckovic2017graph}, GIN \cite{xu2018powerful}, GraphSAGE \cite{hamilton2017inductive}, GraphMAE \cite{hou2022graphmae} and GDGIN \cite{kong2022geodesic}; \textbf{(\uppercase\expandafter{\romannumeral2}) Graph Pooling method}: GMT \cite{baek2021accurate}, TopKPool \cite{gao2019graph}, SAGPool \cite{lee2019self}, DiffPool \cite{ying2018hierarchical}, MinCutPool \cite{bianchi2020spectral}, SUGAR \cite{sun2021sugar}, GINICL \cite{zhao2024twist} and TIP \cite{ying2024boosting}. 

Based on the results presented in Table \ref{tab:updated_accuracy_percentages_new}, our comparison highlights the following key advantages of TPAGP:
\textbf{Superior Performance Across Datasets:} TPAGP achieves the highest accuracy on all benchmark datasets, significantly surpassing existing baselines. Notably, on challenging datasets such as DD and PTC\_MR, TPAGP outperforms the best alternative models by over 3\%, demonstrating its robustness in diverse graph structures.
\textbf{Adaptive Graph Coarsening:} Unlike fixed pooling strategies such as TopKPool and SAGPool, TPAGP dynamically refines its granularity based on the intrinsic graph structure. This adaptability ensures retention of crucial node and edge information, leading to more accurate graph representations.
\textbf{Robustness in Diverse Domains:} While models such as SAGPool and DiffPool exhibit performance fluctuations across datasets, TPAGP maintains a consistently high accuracy, showcasing its ability to generalize across different types of graphs, from molecular datasets (MUTAG, BZR) to social network datasets (IMDB-MULTI).
\textbf{Enhanced Representation Learning:} The integration of hierarchical granular-ball structures within TPAGP enables better feature propagation and aggregation, mitigating over-smoothing issues observed in models like GraphSAGE and GraphMAE.

\renewcommand{\arraystretch}{1.01}
\begin{table*}[tb!]
 \caption{Test accuracy (in \%) on the benchmark datasets. The bold numbers represent the improvement of our model over baselines is statistically significant with p-value < 0.01.}
 \label{tab:updated_accuracy_percentages_new}
 \begin{tabularx}{\linewidth}{l >{\centering\arraybackslash}X >{\centering\arraybackslash}X >{\centering\arraybackslash}X >{\centering\arraybackslash}X >{\centering\arraybackslash}X >{\centering\arraybackslash}X >{\centering\arraybackslash}X }
    \hline
    Models & MUTAG & MSRC\_9 & BZR & DD & PTC\_MR & IMDB & AVG \\
    \hline    
        GCN         & 76.60  & 94.64  & 85.29  & 73.9  & 67.44  & 34.93  & 72.13  \\
        GAT         & 74.47 & 92.86  & 83.33  & 59.66 & 59.3   & 36.27  & 67.65  \\
        GIN         & 84.03 & 95.48  & 84.31  & 79.7  & 58.13  & 48.8   & 75.08  \\
        GraphSAGE   & 82.98 & 91.07  & 79.41  & 74.24 & 61.63  & 32.00  & 70.22  \\
        GraphMAE    & 73.19 & 93.7   & 78.77  & 60.27 & 55.53  & 43.2   & 67.44  \\
        GMT         & 80.85 & 10.71  & 86.27  & 75.81 & 54.65  & 41.33  & 58.27  \\
        GDGIN       & 85.1  & 26.78  & 86.27  & 76.95 & 63.95  & 46.39  & 64.24  \\
    \hline
        TopKPool    & 86.11 & 93.18  & 90.00  & 76.92 & 64.70  & 44.66  & 75.93  \\
        SAGPool     & 83.78 & 47.73  & 81.48  & 74.04 & 69.12  & 42.00  & 66.36  \\
        DiffPool    & 88.89 & 95.45  & 82.50  & 79.31 & 61.76  & 46.67  & 75.76  \\
        MinCutPool  & 72.97 & 81.82  & 83.95  & 75.32 & 63.24  & 33.67  & 68.50  \\
        SUGAR       & 89.25 & 94.06  & 81.48  & 81.10 & 65.12  & 51.67  & 77.11  \\
        GINICL      & 88.64 & 95.45  & 87.10  & 77.97 & 66.28  & 51.34  & 77.80  \\
        TIP         & 86.84 & 95.56  & 85.19  & 77.12 & 65.22  & 52.33  & 77.04  \\
    \hline
    \textbf{TPAGP}  & \textbf{89.47} & \textbf{97.78} & \textbf{97.56} & \textbf{83.05} & \textbf{76.47} & \textbf{53.33} & \textbf{82.94} \\
    \hline
 \end{tabularx}
\end{table*}

\subsection{Ablation Studies}
To investigate \textbf{RQ2}, we conduct an ablation study using three configurations of our proposed TPAGP model to evaluate the impact of its key components. The configurations are: \textbf{(1) TPAGP (full model)}, \textbf{(2) - w/o \textsl{Granular-Ball Graph Pooling}}, which removes the granular-ball generation and multi-granularity interaction mechanisms, and \textbf{(3) - w/o \textsl{Multi-granularity Graph Network Model}}, which excludes all TPAGP-specific components, retaining only basic graph processing techniques.  

We evaluate these configurations on multiple datasets, and the results are summarized in Table \ref{tab: Ablation Studies}. Our analysis highlights the following key observations:  

\renewcommand{\arraystretch}{1.35}
\begin{table*}[htbp]
  \centering
  \caption{Results of the Ablation Study.}
  \label{tab: Ablation Studies}
  \begin{tabularx}{\textwidth}{l >{\centering\arraybackslash}X >{\centering\arraybackslash}X >{\centering\arraybackslash}X >{\centering\arraybackslash}X >{\centering\arraybackslash}X >{\centering\arraybackslash}X >{\centering\arraybackslash}X}
    \toprule
    Models & MSRC\_9 & BZR & DD & PTC\_MR  & AVG \\
    \midrule
    ours  & 97.78 & 97.56 & 83.05 & 76.47  & 88.71 \\
    - w/o \textsl{Granular-Ball Graph Pooling}  & 95.45 & 92.68 & 80.51 & 70.58 & 84.80 \\
    - w/o \textsl{Multi-granularity Graph Network Model}  & 95.08 & 85.91 & 61.43 & 65.59 & 77.00 \\
    \bottomrule
  \end{tabularx}
\end{table*}

The comparison between TPAGP and - w/o \textsl{Granular-Ball Graph Pooling} reveals a significant performance drop when these mechanisms are removed. This demonstrates that Granular-Ball Graph Pooling is critical for capturing and integrating graph structure and node features, enabling the model to effectively preserve both local and global patterns.  

The - w/o \textsl{Multi-granularity Graph Network Model} configuration, which excludes the overall TPAGP design, shows particularly poor performance, especially in terms of AUC. This underscores the essential role of the complete TPAGP framework in leveraging graph topology and feature interactions to achieve robust and effective representation learning.  

These results confirm that the granular-ball generation and multi-granularity interaction mechanisms, along with the overall TPAGP architecture, are essential for achieving superior performance in graph representation learning tasks.

\subsection{Hyper-parameter Sensitivity Analysis}
\begin{wrapfigure}{r}{0.4\textwidth} 
  \includegraphics[width=\linewidth]{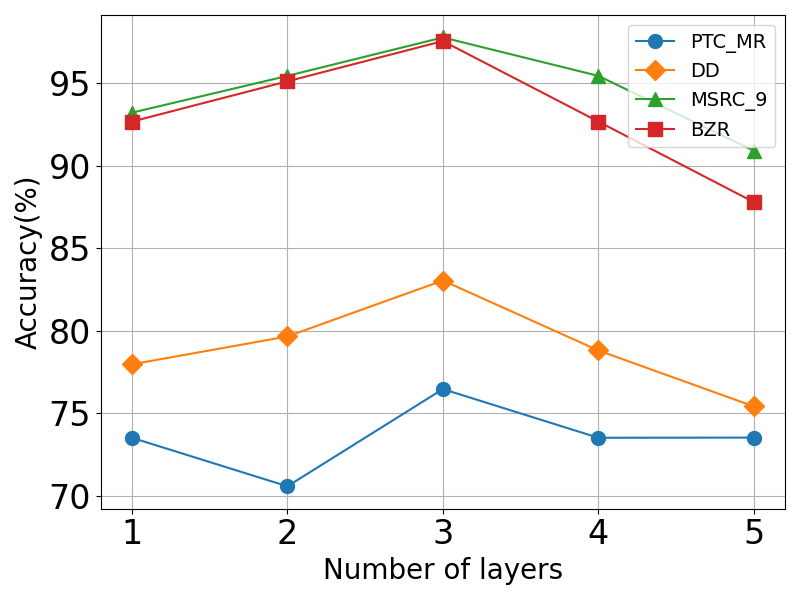}
  \caption{Hyper-parameter}
  \label{fig:hyper-param-sensitivity-analysis}
\end{wrapfigure}

To address \textbf{RQ3}, we perform an in-depth examination of the critical hyper-parameters affecting the TPAGP model's performance, with emphasis on the number of GNN layers. The number of GNN layers is a critical hyperparameter, as it controls the depth of feature aggregation and thus influences the effectiveness of topology-preserving adaptive pooling. Our findings reveal that appropriately tuning this parameter markedly enhances the model's overall effectiveness, enabling further refinement of TPAGP. The corresponding results are illustrated in Fig.\ref{fig:hyper-param-sensitivity-analysis}.

The results indicate that increasing the GNN layer depth from 2 to 3 consistently improves performance across all datasets. This suggests that TPAGP benefits from deeper feature propagation, which enhances its ability to exploit hierarchical structural information generated by topology-preserving pooling. However, increasing the depth to 4 layers results in performance degradation, likely due to over-smoothing, where node representations become indistinguishable after excessive aggregation. This trend aligns with TPAGP’s core design, which relies on adaptive pooling guided by local topology—over-smoothing weakens the discriminative power of the representations used in granular-ball quality assessment and pooling decisions.
Therefore, using 3 GNN layers provides the most favorable balance between expressive power and topological fidelity within the TPAGP framework.

\subsection{Case Study}
\begin{figure*}[h]
    \centering
    \begin{subfigure}[b]{0.3\textwidth}
        \centering
        \includegraphics[width=\linewidth]{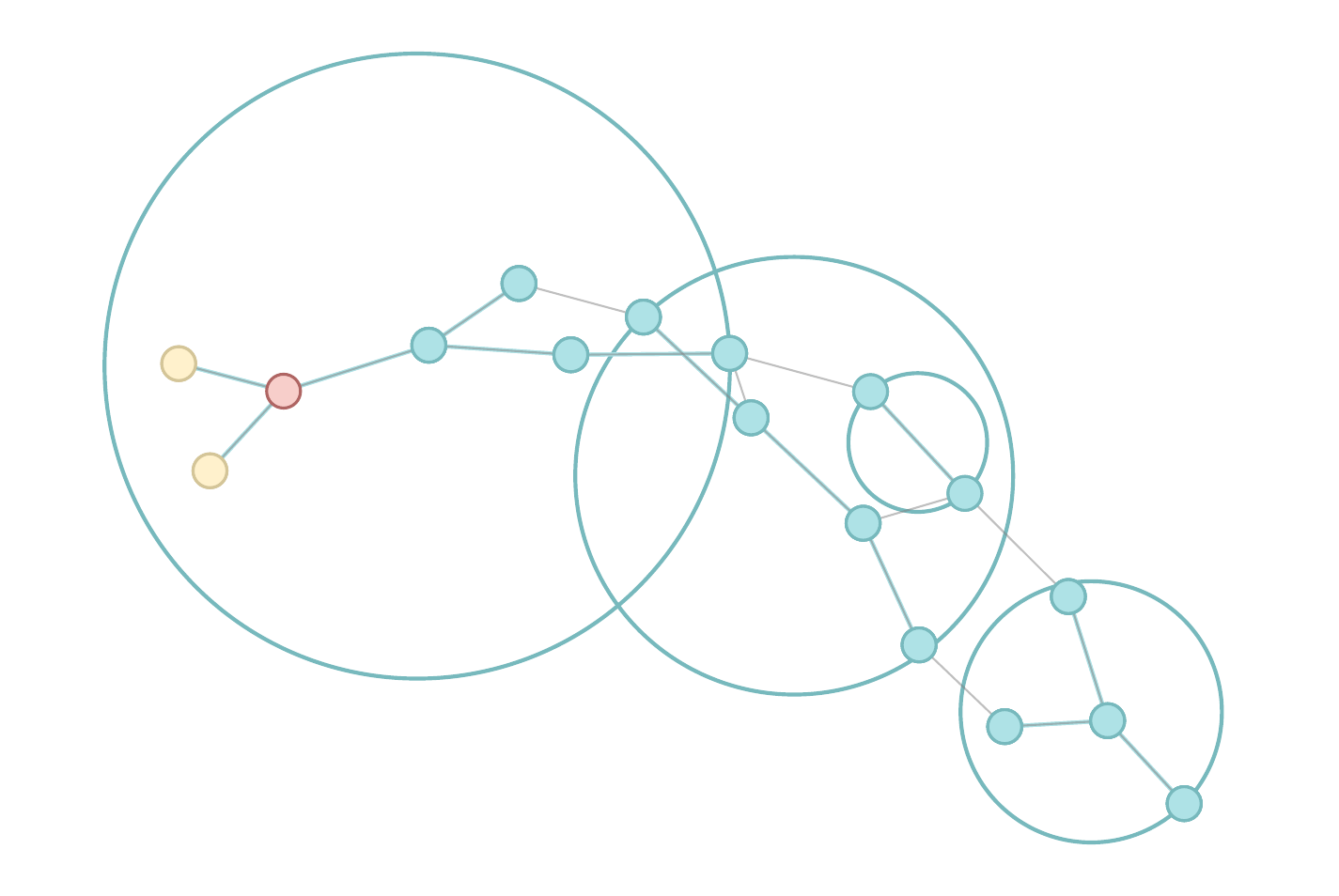}
         \caption{MUTAG}
    \end{subfigure}
    \hfill
    \begin{subfigure}[b]{0.3\textwidth}
        \centering
        \includegraphics[width=\linewidth]{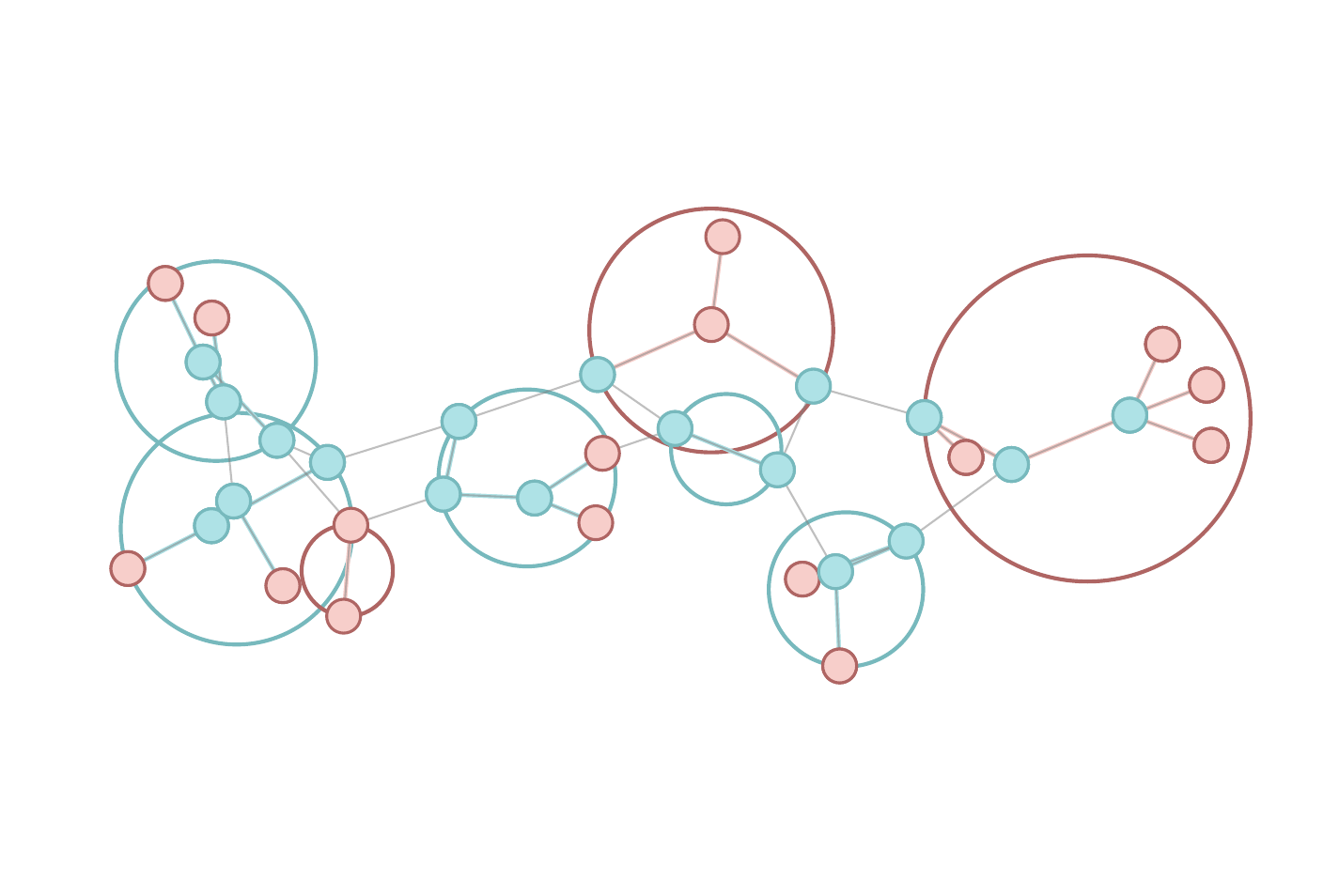}
        \caption{BZR}
    \end{subfigure}
    \hfill
    \begin{subfigure}[b]{0.3\textwidth}
        \centering
        \includegraphics[width=\linewidth]{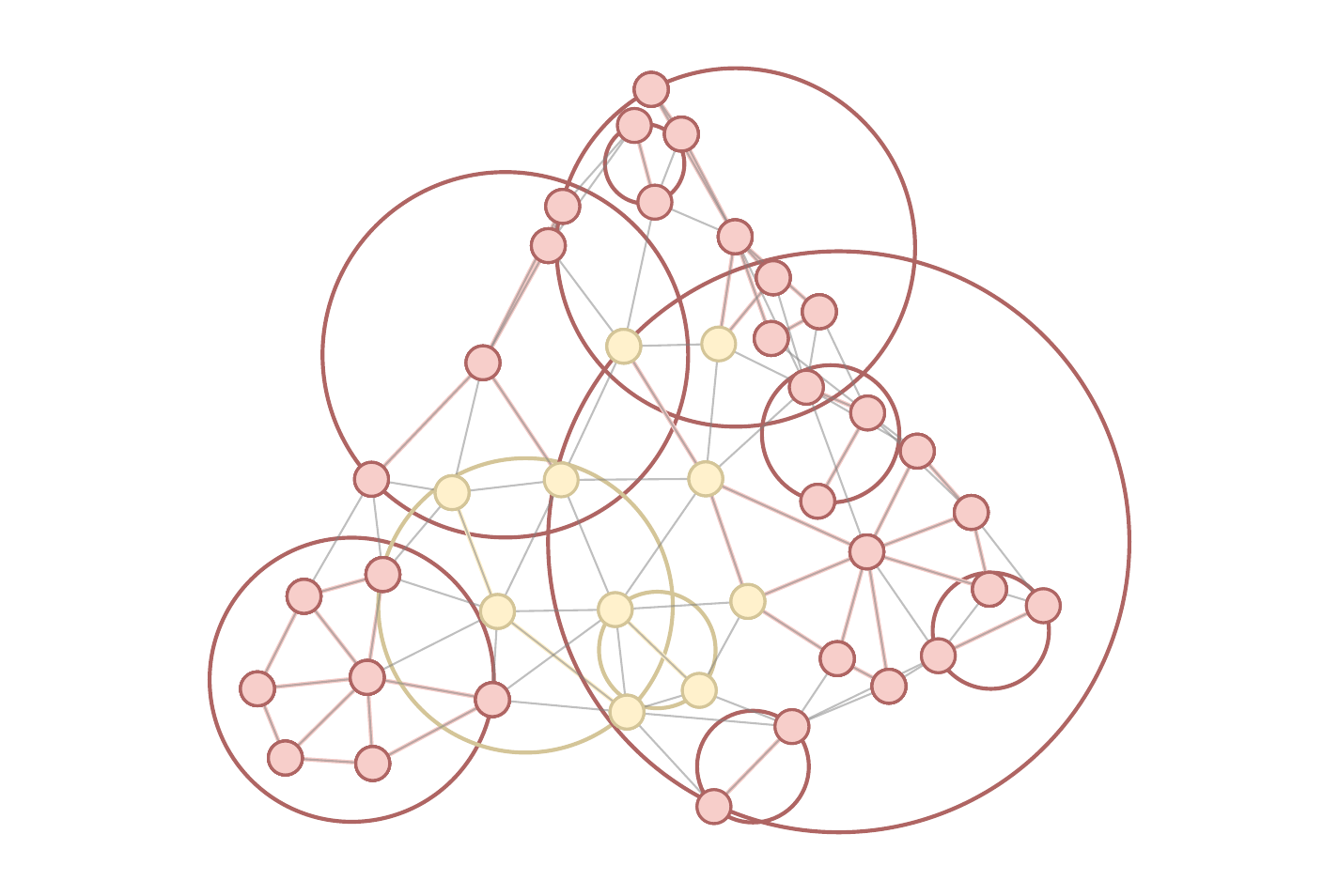}
        \caption{MSRC\_9}
    \end{subfigure}
    \caption{Visualization of TPAGP in Different Datasets.}
    \label{fig:Case}
\end{figure*}
To answer \textbf{RQ4}, this section examines the application of the TPAGP method to the datasets, focusing on its impact, domain relevance, and pooling effects, as shown in Figure \ref{fig:Case}. In the figure, nodes of different colors represent distinct categories, while the circles represent the granular balls generated by the TPAGP method. Nodes within each granular ball are aggregated into a pooled super node. TPAGP combines local and global topological features to generate adaptive granular balls for multi-level pooling, facilitating feature extraction at varying granularities. The adaptive splitting conditions, based on subgraph topological features, ensure that the formed granular balls align with the data's intrinsic structure and that each granular ball's granularity is optimal, thereby enhancing pooling effectiveness.

\section{Conclusion}
This paper presents the TPAGP method, a novel adaptive solution for graph pooling in graph neural networks. By dynamically adjusting the granularity of granular-balls, TPAGP overcomes the limitations of fixed-granularity methods, effectively adapting to diverse graph characteristics. Extensive experiments validate its superior performance in graph classification tasks compared to state-of-the-art methods. Future work may explore extending TPAGP to other graph-related tasks and optimizing its computational efficiency for large-scale graph datasets.

{
\small
\bibliographystyle{plain} 
\bibliography{cite}
}

\newpage
\appendix


\section{Dataset Introduction}
\label{appendix:dataset}
\begin{itemize}
    \item \textbf{MUTAG} is a small molecular graph dataset used for predicting the mutagenicity of molecules.  

    \item \textbf{MSRC\_9} is a semantic graph dataset for image segmentation tasks, aimed at predicting region categories.  

    \item \textbf{BZR} focuses on studying the interactions of bioactive molecules.  

    \item \textbf{DD} is a large dataset containing protein structures for biological tasks.  

    \item \textbf{PTC\_MR} is a molecular graph dataset used for predicting carcinogenicity in rodents.  

    \item \textbf{IMDB-MULTI} is a social network dataset designed for movie genre classification based on graph structures.  
\end{itemize}

\end{document}